\documentclass[letterpaper]{article} 
\usepackage[submission]{aaai2026}  
\usepackage{times}  
\usepackage{helvet}  
\usepackage{courier}  
\usepackage[hyphens]{url}  
\usepackage{graphicx} 
\usepackage{natbib}  
\usepackage{caption} 
\usepackage{algorithm}
\usepackage{algorithmic}

\usepackage{graphicx}
\usepackage{amsmath}
\usepackage{amsthm}
\usepackage{booktabs}
\usepackage{algorithm}
\usepackage{algorithmic}
\usepackage[switch]{lineno}

\usepackage{multirow} 
\usepackage{color}

\usepackage{newfloat}
\usepackage{listings}
\DeclareCaptionStyle{ruled}{labelfont=normalfont,labelsep=colon,strut=off} 
\floatstyle{ruled}
\newfloat{listing}{tb}{lst}{}
\floatname{listing}{Listing}
\title{
GoAI: Enhancing AI Students’ Learning Paths and Idea Generation via Graph of AI Ideas
}
\author{
    Written by AAAI Press Staff\textsuperscript{\rm 1}\thanks{With help from the AAAI Publications Committee.}\\
    AAAI Style Contributions by Pater Patel Schneider,
    Sunil Issar,\\
    J. Scott Penberthy,
    George Ferguson,
    Hans Guesgen,
    Francisco Cruz\equalcontrib,
    Marc Pujol-Gonzalez\equalcontrib
}
\affiliations{
    \textsuperscript{\rm 1}Association for the Advancement of Artificial Intelligence\\

    1101 Pennsylvania Ave, NW Suite 300\\
    Washington, DC 20004 USA\\
    proceedings-questions@aaai.org
}

\title{GoAI: Enhancing AI Students’ Learning Paths and Idea Generation via Graph of AI Ideas}
\author {
    Xian Gao\textsuperscript{\rm 1},
    Zongyun Zhang\textsuperscript{\rm 1},
    Ting Liu\textsuperscript{\rm 1},
    Yuzhuo Fu\textsuperscript{\rm 1} \\
    \textsuperscript{1}Shanghai Jiao Tong University
}

\usepackage{bibentry}

\begin{document}

\maketitle

\begin{abstract}

With the rapid advancement of artificial intelligence technology, AI students are confronted with a significant "information-to-innovation" gap: they must navigate through the rapidly expanding body of literature, trace the development of a specific research field, and synthesize various techniques into feasible innovative concepts. An additional critical step for students is to identify the necessary prerequisite knowledge and learning paths. Although many approaches based on large language models (LLMs) can summarize the content of papers and trace the development of a field through citations, these methods often overlook the prerequisite knowledge involved in the papers and the rich semantic information embedded in the citation relationships between papers. Such information reveals how methods are interrelated, built upon, extended, or challenged. To address these limitations, we propose GoAI, a tool for constructing educational knowledge graphs from AI research papers that leverages these graphs to plan personalized learning paths and support creative ideation. The nodes in the knowledge graph we have built include papers and the prerequisite knowledge, such as concepts, skills, and tools, that they involve; the edges record the semantic information of citations (e.g., baseline, extension, comparison). When a student queries a specific paper, a beam search-based path search method can trace the current development trends of the field from the queried paper and plan a learning path toward cutting-edge objectives. The integrated Idea Studio guides students to clarify problem statements, compare alternative designs, and provide formative feedback on novelty, clarity, feasibility, and alignment with learning objectives. Test results, including those on learning gains, creativity quality, and cognitive load, indicate that graph-based learning path planning, creativity generation, and structured feedback can enhance students' ability to summarize frontiers, identify key technologies, and generate feasible concepts.

\end{abstract}

\section{Introduction}


Generating feasible and innovative ideas lies at the heart of AI education. Traditionally, this process requires students to read academic papers in the AI domain, trace research trends, identify and master the prerequisite knowledge embedded within those works, and, based on the prevailing research directions and cutting-edge advancements, formulate novel ideas. However, in the rapidly evolving field of artificial intelligence—characterized by an explosive growth in publications and increasingly interdisciplinary research—students are often overwhelmed by the sheer volume and intricate citation networks of scientific literature. As a result, they must devote significant time and effort to navigate this landscape. The insufficiency of comprehensive literature reviews further exacerbates the difficulty, making it challenging for students to gain a holistic understanding of a topic or to chart a personalized learning pathway from their current knowledge base toward the generation of innovative ideas.


Although LLM-based academic assistants are capable of answering questions, summarizing papers, and even generating novel ideas \cite{wangSciPIPLLMbasedScientific2024, wangSciMONScientificInspiration2024, liChainIdeasRevolutionizing2024}, they typically provide only a linear narrative of research development, which presents three fundamental limitations: (i) they overlook pedagogical prerequisites—namely, what foundational knowledge must be acquired first and how the difficulty should progress—rendering the output less practical for direct student use; (ii) they fail to capture the semantic roles of citations, such as whether a cited work serves as a baseline to be surpassed, an extensible foundation, or a critique to be addressed; and (iii) their linear depiction oversimplifies the intricate structure of academic development, which is better represented as a graph where the interconnections among scholarly works form a complex knowledge network. These limitations hinder both learning efficiency and the quality of idea generation. In essence, the intellectual landscape that AI students must navigate is inherently graph-structured: research papers and their corresponding ideas or proposed methods serve as nodes; citation relationships—each carrying functional semantic roles—form the edges; and the concepts and skills embedded in these works create prerequisite chains. A knowledge graph perspective unveils this structure, enabling algorithmic planning of personalized learning pathways and revealing “innovation opportunities”—points where methods may be integrated, extended, or contrasted to inspire creative research directions.


To address these limitations, we introduce GoAI, a novel framework that leverages knowledge graphs and large language models to analyze the relationships among AI research ideas and generate personalized learning pathways for students. First, we construct a research-centric knowledge graph for AI education by extracting paper entities and semantically annotated citation relationships. These semantic citation links encode the functional roles between scholarly works—such as baseline, extension, contrast/ablation, or critique/response—while the paper entities encompass not only the publications themselves but also the methods proposed, concepts involved, and prerequisite knowledge or skills required. Second, given a student’s target topic or research interest, we retrieve the most relevant papers from the graph and employ beam search guided by citation semantics to generate multiple development trajectories, summarizing the evolution of research in the field. The search process prioritizes citation types most conducive to achieving the learning objective (e.g., understanding baselines before exploring extensions) and organizes the resulting papers into a progressive curriculum based on the complexity of their prerequisite concepts. Finally, within the integrated Idea Studio, students synthesize their acquired knowledge into project ideas , receiving formative feedback from a Chain-of-Thought (CoT) reviewer that evaluates the novelty, strengths, and potential limitations of their proposals.


The main contributions of this study are summarized as follows
\begin{itemize}
    \item We propose GoAI, a knowledge graph for AI education that is centered around research papers and enriched with metadata such as concepts and skills, while preserving semantically meaningful citation labels. This integrated structure bridges pedagogy and scholarly research, supporting both learning and creative ideation.

    \item Building upon GoAI, we introduce a learning path planning method in which large language models dynamically traverse the AI graph to analyze research trends and generate personalized learning trajectories.

    \item We design an Idea Studio that assists students in refining their research ideas based on identified trends, and enhances the transparency of the evaluation process through structured assessments. 

    \item Experimental results demonstrate that the GoAI framework effectively reduces students’ cognitive load, leading to measurable improvements in both learning outcomes and the quality of generated ideas.
    
\end{itemize}
\section{Related Works}

\subsection{Knowledge Graph in Education}

In the field of education, knowledge graphs have made notable progress in relationship modeling and practical applications, with increasing maturity in large-scale resource integration and evaluation ecosystems \cite{abu-salihSystematicLiteratureReview2024a, quSurveyKnowledgeGraph2024a}. In terms of graph construction, EDUKG \cite{zhaoEDUKGHeterogeneousSustainable2022}, designed for K–12 education, integrates textbooks and educational resources through fine-grained ontologies and sustainable maintenance mechanisms, laying a foundation for unified modeling across interdisciplinary knowledge and resources. Other efforts, such as interactive annotation pipelines leveraging human–AI collaboration \cite{aytekinACEAIAssistedConstruction2024}, have significantly reduced expert annotation workload. Automated extraction and fusion frameworks for multi-source educational knowledge graphs have also been explored and validated in domain-specific contexts \cite{liMultiSourceEducationKnowledge2023}. On the application side, early representation learning methods within MOOCs \cite{panPrerequisiteRelationLearning2017} and recent enhancements using directed GNNs to capture richer dependencies \cite{quConceptPrerequisiteRelation2024} highlight sustained advancements in learning path planning and curriculum dependency modeling. Knowledge graphs have been integrated with semantic encoders such as GNNs and SBERT to enable interpretable and fine-grained recommendation systems on MOOC platforms \cite{alatrashConceptGCNKnowledgeConcept2024}. EDUKG has further been employed to provide factual context for LLMs to generate auditable and explainable learning recommendations \cite{abu-rasheedKnowledgeGraphsContext2024a}, and graph-based retrieval-augmented generation (GraphRAG) approaches tailored to educational path planning have begun to emerge \cite{abdelmagiedLeveragingGraphRetrievalAugmented2025}. However, these studies have largely concentrated on the integration of static knowledge, with limited attention to leveraging knowledge graphs for the synthesis of academic literature and the facilitation of student-driven innovation.

\subsection{AI-based Scientific Discovery}

Artificial intelligence has become a powerful tool for advancing research processes, reshaping scientific discovery by reducing the time needed for experimental iterations, and assisting researchers in various capacities \cite{yuG2TLLMGraphtoTreeText2024,abramsonAccurateStructurePrediction2024,haanAstroMLab3Achieving2024,tanChatGPTMedicineProspects2024}. The development of large language models, with their advanced natural language processing capabilities and ability to integrate interdisciplinary information, enables more efficient literature analysis and the generation of novel hypotheses \cite{kumarCanLargeLanguage2024,siCanLLMsGenerate2024}. Prior research has enhanced novelty by iteratively comparing AI-generated ideas with existing literature \cite{wangSciMONScientificInspiration2024}, while other studies have focused on linking academic knowledge through core papers, expanding understanding via academic graphs and entity-centered knowledge stores \cite{baekResearchAgentIterativeResearch2024,wangSciPIPLLMbasedScientific2024,liChainIdeasRevolutionizing2024}. Some studies use multi-agent systems based on LLMs to foster the generation of research ideas in autonomous scientific discovery \cite{suTwoHeadsAre2024}. In this paper, we adopt a paradigm for generating research ideas based on academic graphs derived from core papers. Unlike previous works, we represent citation relationships between academic papers and summarize development trends through a knowledge graph, rather than through research trend chains. Furthermore, we introduce the concept of constructing a knowledge graph that incorporates semantic information and citation importance, providing richer information for analyzing relationships between papers.

\subsection{Graph-augmented LLMs}
The integration of LLMs with knowledge graphs, which store information in natural language, is an emerging research area aimed at fully utilizing the explicit and updatable external knowledge within structured graphs to enhance the capabilities of large language models \cite{panUnifyingLargeLanguage2024,huangCanGNNBe2024}, particularly in tasks such as reasoning \cite{luoReasoningGraphsFaithful2024,luoGraphconstrainedReasoningFaithful2024} and rule completion \cite{luoChatRuleMiningLogical2024}
. To input the structured knowledge from knowledge graphs into large language models, some methods transcribe entity-relation-entity triples from the knowledge graph into natural language text that is understandable by the language model, which is then provided as a prompt to the LLM \cite{sunThinkonGraphDeepResponsible2024}, similar to retrieval-augmented generation. Other methods directly input graph information processed by approaches such as Graph Neural Networks (GNNs) into the large language model, utilizing the implicit graph information for enhancement \cite{huangCanGNNBe2024,mavromatisGNNRAGGraphNeural2024}. In this paper, since the literature used to construct the graph is already in natural language and the relations within the graph contain semantic information in natural language, we adopt the method of transcribing the graph into textual prompts for input into the large language model to achieve the collaboration between the graph and the model.

\begin{figure*}[!t]
    \centering
    \includegraphics[width=0.95\linewidth]{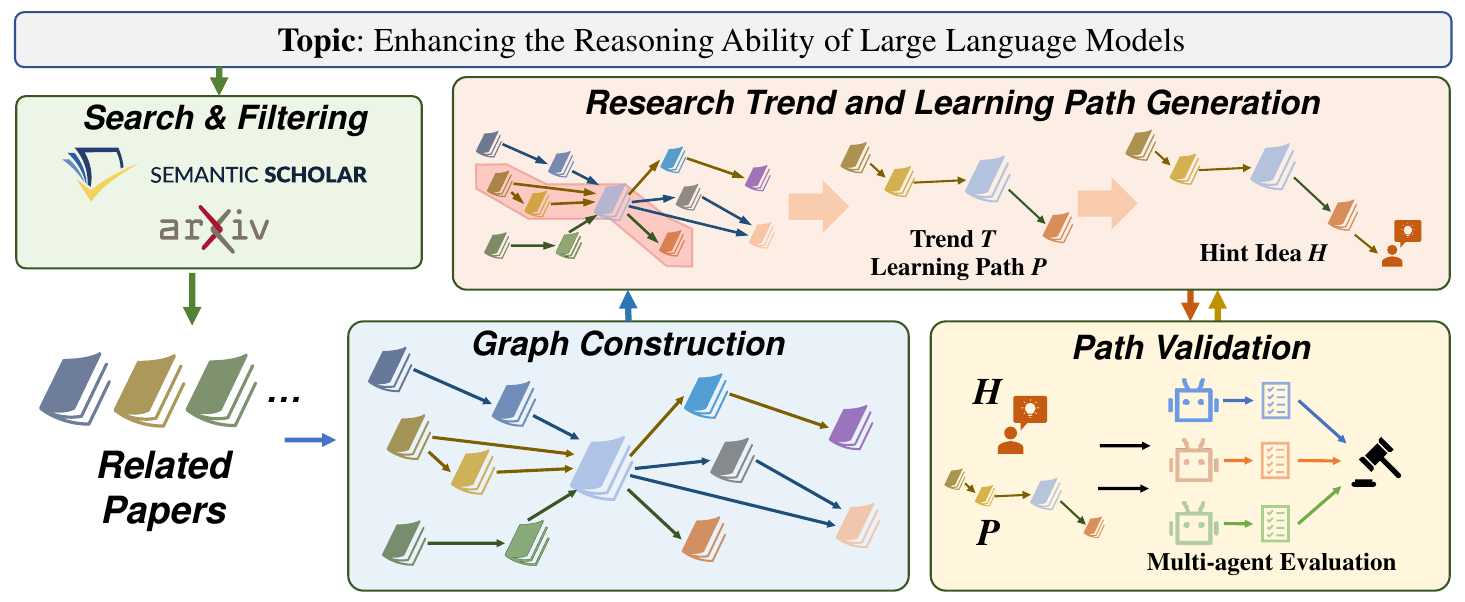}
    \caption{The framework of GoAI. The framework consists of four stages: (1) Literature Search and Filtering, (2) GoAI Graph Construction, (3) Path Generation, and (4) Path Validation.}
    \label{fig:framework}
\end{figure*}

\section{Methodology}
\subsection{GoAI Framework}

In this section, we present a detailed overview of the GoAI framework. As illustrated in Figure \ref{fig:framework}, the framework comprises four stages: (1) literature search and filtering, (2) construction of the GoAI knowledge graph, (3) path generation, including generating research trends and learning pathways, and (4) validation of pathway coherence. Given a research topic, GoAI first searches online academic repositories for relevant literature and identifies key reference papers based on their ranking. It then expands the knowledge graph bi-directionally by incorporating papers connected via citation links to these key references. For each paper, the core idea is extracted to reflect the research trajectory of the field, while the embedded AI-specific prerequisite knowledge is gathered to support student learning. Subsequently, an LLM-as-agent analyzes the graph to synthesize the evolving research trends and constructs a personalized, prerequisite-aligned learning path. Based on this trajectory, the model generates ``Hint Ideas” to inspire innovation. Finally, a multi-agent evaluation system assesses these Hint Ideas and learning paths, validating the coherence of the proposed pathways through iterative refinement and integration, thereby ensuring the plausibility and academic soundness of the generated research progression and learning paths.

\subsection{Literature Search and Filtering}

We use the Semantic Scholar API\footnote{https://www.semanticscholar.org/product/api} and Arxiv API\footnote{https://info.arxiv.org/help/api/index.html} to search for papers in the online paper database. Given an initial research topic, we retrieve the top-ranked papers, which we refer to as key references, forming the foundation for building the graph. Starting from the corresponding key reference papers, GoAI constructs the graph by extending in both forward and backward directions based on citation relationships. The backward extension is derived from the reference list of the key reference, tracing the origins of the paper's ideas. The forward extension is achieved by exploring which papers in the literature cite the key reference paper, representing the subsequent development of the idea. 

For the papers in the backward extension, we rank them based on their cosine similarity to the initial research topic and the abstract of the key reference paper. The top $K$ papers are selected and added to the graph, representing one step of the process. For each new paper added to the graph, the same rule is applied, extending backward based on its reference list until the number of backward steps reaches the set upper limit $N$. A similar procedure is applied in the forward direction, extending $N$ steps. If no references relevant to the original research topic are found during the extension process, GoAI terminates the search. Under ideal conditions, the number of papers retrieved is $O(2K^N)$.

\subsection{Graph Construction}
\subsubsection{GoAI Graph Overview}
The GoAI graph is used to represent the relationships between papers that have citation relationships, including the citing and cited papers, the importance of the citation (measured by the position where the reference appears in the original text), and the semantic information contained in the citation. The general representation of the GoAI graph is a structured	$<$\textit{paper1, citation position, citation semantics, paper2}$>$ quadruple. The citation position and citation semantics together form the relation field in traditional knowledge graphs, which is used for subsequent analysis of the graph. To construct the GoAI graph, it is necessary to separately obtain the papers as entities and the citation positions and citation semantics as relations.

\subsubsection{Relation Definition and Extraction}

After determining the entities to be included in the graph, we proceed with constructing the relations between these entities. For the relationships between different papers, we define the citation relationships between papers as one of the following five semantic categories:
\begin{itemize}
    \item \textbf{Based on and Extension (B\&E)}: The paper is based on, extends, applies, or is inspired by the cited paper, or generalizes its theories and methods.
    \item \textbf{Support and Supplement (S\&S)}: The paper supports the cited paper’s work through citation, reuse, supplementation, or indirect connection.
    \item \textbf{Contrast and Alternative (C\&A)}: The paper compares itself with the cited paper, proposes alternatives, or summarizes its content from a different perspective.
    \item \textbf{Question and Refutation (Q\&A)}: The paper questions, corrects, or refutes the content of the cited paper.
    \item \textbf{Simple Mention or Irrelevant (M/I)}: The paper merely mentions the cited paper, or there is little to no direct relevance between the two.
\end{itemize}

We use the SciPDF Parser\footnote{https://github.com/titipata/scipdf\_parser} to parse the paragraphs from the paper documents, prompting the LLM to extract the cited references from the original paragraphs and determine which of the five semantic relationships best describes the citation.

\subsection{Path Generation}
The process of generating paths from the graph is divided into two steps: graph exploration and path generation. Figure \ref{fig:graph} illustrates an example workflow of graph exploration and path generation.

    
        

\subsubsection{Graph Exploration}

In the graph, the entities are papers, and the relations are represented as pairs of citation positions and citation semantics. At the beginning of the $D$-th iteration, each path consists of $2D-1$ quadruples, i.e., $ p = \{ ( e^d_s, r^d, e^d_o ) \}^{2D-1}_{d=1} $, where $e_s $ and $e_o$  represent the subject and object entities, respectively, and $r$ denotes the specific relation between them. The pairs $(e^d_s, r^d, e^d_o)$ and $(e^{d+1}_s, r^{d+1}, e^{d+1}_o)$ are interconnected. The sets of endpoint entities and relations of all paths in the path set $P$ are denoted as  $E^{D-1}=\{e^{d-1}_1,e^{d-1}_2,\cdots,e^{d-1}_N\}$  and  $R^{D-1}=\{r^{d-1}_1,r^{d-1}_2,\cdots,r^{d-1}_N\}$, respectively.


The exploration phase of the $D$-th iteration aims to utilize the LLM to identify the top-N most relevant entities  $E^{D}$  from the neighboring entities of the current top-N endpoint entity set $E^{D-1}$, and use $E^D$ to extend the top-N paths in the path set $P$. The endpoint entities here can be either the head or tail entities of the path, corresponding to backward and forward extensions, respectively. To address the complexity of processing a large number of neighboring entities using the LLM, inspired by the approach of ToG \cite{sunThinkonGraphDeepResponsible2024}, this paper implements a two-step exploration strategy: first, explore the important relations, and then use the selected relations to guide entity exploration.

\begin{figure*}[!t]
    \centering
    \includegraphics[width=0.95\linewidth]{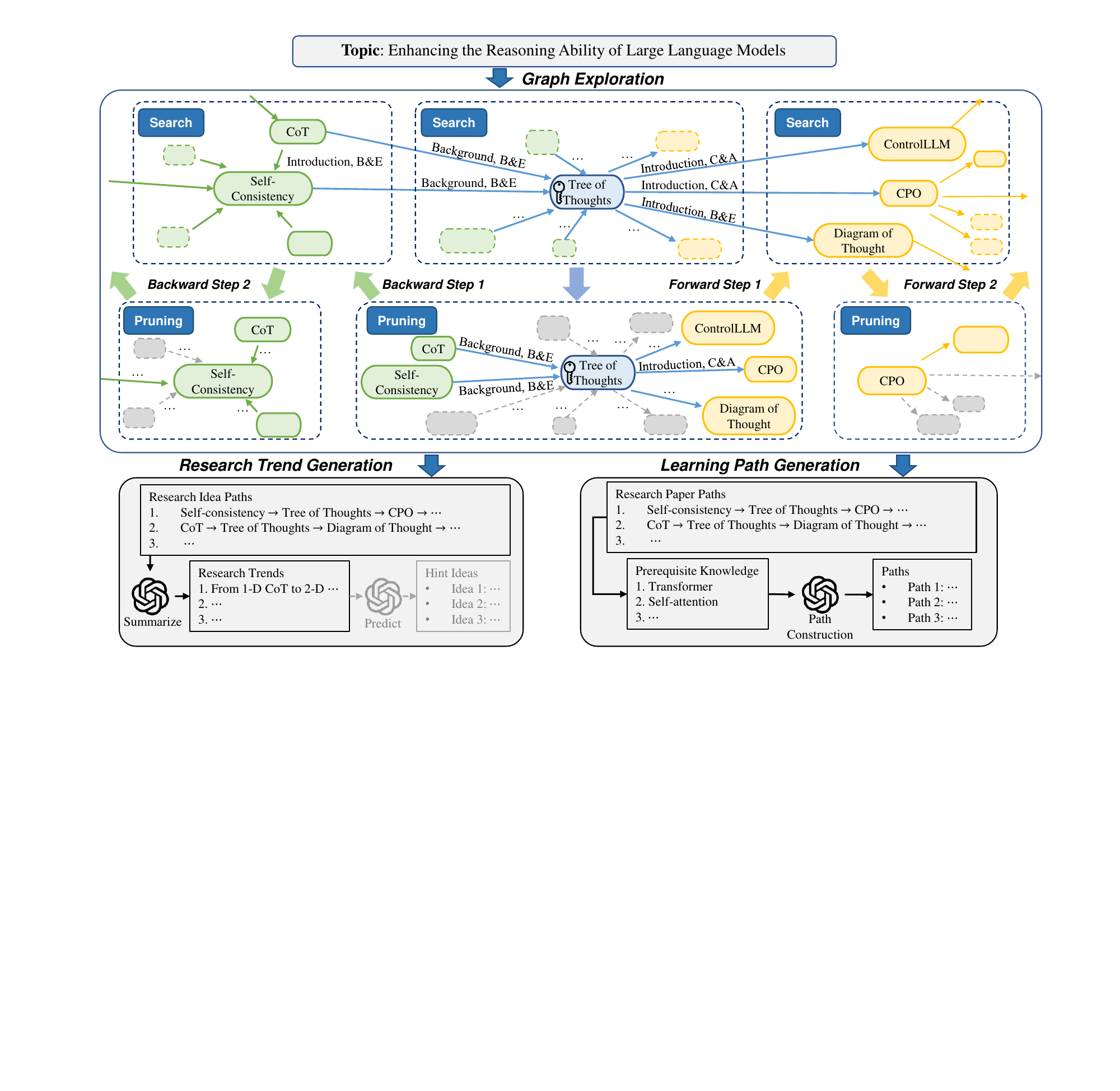}
    \caption{An example workflow of graph exploration and idea generation. The paper ``Tree of Thoughts" is the key reference. A more detailed sub-graph example can be found in our supplementary materials.}
    \label{fig:graph}
\end{figure*}

\textbf{Relation Exploration} The relation exploration is a beam search process with a depth of 1 and a width of $N$, from  $E^{D-1}$  to  $R^D$ . The entire process can be broken down into two steps: search and pruning. We use the GoAI Agent, an LLM-based agent, to automatically complete this process.

\begin{itemize}
    \item \textbf{Search Step} At the beginning of the $D$-th iteration, the relation exploration phase first associates each path  $p_n$  with the candidate relations  $R^D_{cand,n}$  that are related to the endpoint entity $e_n^{D-1}$. These relations are then aggregated into $ R^D_{cand}$. In the case of Figure \ref{fig:graph}, $E^1=$ \{\textbf{Tree of Thoughts}\} and $R^1$ denotes the set of all relations linked to \textbf{Tree of Thoughts} inwards or outwards.
    

    \item \textbf{Pruning Step} Once the candidate relation set $ R^D_{cand} $ and the extended candidate paths $ P_{cand} $ are obtained from the relation search, the LLM-based GoAI Agent can be used to select the top-N new paths $ P $ that have the relation $ R^D $ as the endpoint from $ P_{cand} $, based on the textual information of the query and the candidate relations $ R^D_{cand} $. The pruning process can be performed by prompting the LLM, allowing the GoAI Agent to adapt well to different graphs without any training cost.  As shown in Figure \ref{fig:graph}, the LLM selects relations {(Background, B\&E), (Introduction, C\&A), (Introduction, B\&E)} out from all relations linked to the entity \textbf{Tree of Thoughts} in the first iteration. Since \textbf{Tree of Thoughts} is the only topic entity, the candidate paths are updated as \{(Tree of Thoughts, (Background, B\&E)), (Tree of Thoughts, (Tree of Thoughts, C\&A)),(Tree of Thoughts, (Introduction, C\&A))\}.

\end{itemize}

\textbf{Entity Exploration} Similar to relation exploration, entity exploration is also a beam search process performed by the LLM from $ R_D $ to $ E_D $, consisting of two steps: Search and Prune.

\begin{itemize}
    \item \textbf{Search Step} Once we obtain the top-N new paths $ P $ and the new set of relations $ R^D $ from the relation exploration, for each relation path $ p_n \in P $, we can search $ (e^{D-1}_n, r^{D}_n, *) $ or $ (*, r^{D}_n, e^{D-1}_n) $ to obtain the candidate entity set $ E^D_{cand,n} $, where $ e^{D-1}_n $ and $ r_n $ represent the endpoint entities and relations of $ p_n $. We can then aggregate $ \{ E^D_{cand,1}, E^D_{cand,2},\cdots, E^D_{cand,n} \} $ into $ E^D_{cand} $ and use the endpoint entities $ E^D_{cand} $ to extend the top-N paths $ P $ to $ P_{cand} $. For the shown case, $E^1_{cand}$ can be represented as \{Self-Consistency, CoT, CPO, Diagram of Thought, ControlLLM, $\cdots$\}.

    \item {Pruning Step} Since each entity in the candidate set $ E^D_{cand} $ is expressed in natural language, we can use the GoAI Agent to select the top-N new paths from $ P_{cand} $, where these paths end with the entity $ E^D $. As shown in Figure \ref{fig:graph}, the current paths $P$ are updated as \{({Tree of Thoughts, (Background, B\&E), Self-Consistency}), ({Tree of Thoughts, (Introduction, C\&A), CoT}),({Tree of Thoughts, (Introduction, C\&A), CPO}),({Tree of Thoughts, (Introduction, B\&E), Diagram of Thought}),({Tree of Thoughts, (Introduction, B\&E), ControlLLM})\}.

\end{itemize}

After performing the above two explorations, we construct the top-N new paths $ P=\{p_n\}_{n=1}^N$, where the length of each path is increased by 2.

\subsubsection{Research Tread and Learning Path Generation with paths}
Upon obtaining the current exploration path $p$, we prompt the LLM to evaluate the developmental trajectory represented by the papers along this path, yielding a synthesized trend $T$. Concurrently, for each generated path, the model is guided to predict potential future research directions. Following this, we further prompt the LLM to articulate the underlying motivation, novelty, and methodological approach, ultimately producing a Hint Idea $H$. This Hint Idea serves a dual purpose: it can be utilized in the subsequent section by the novelty-evaluation agents to evaluate candidate ideas as a proxy for path plausibility, and it also functions as an important stimulus and reference point for fostering student creativity. 


Similarly, once the current path $p$ is determined, the LLM is prompted to extract the AI-related concepts and skills embedded in each paper along the path and to sequence them in ascending order of complexity, thereby constructing a prerequisite-aligned learning path essential for mastering the developmental trajectory of the field.

\subsection{Path Validation}

We construct a multi-agent evaluation system to assess the feasibility of the proposed pathways, encompassing both the research trajectory and the associated learning path. For the learning path, the evaluating agents are prompted to verify the accuracy of the extracted prerequisite knowledge from each paper and to assess the logical sequencing of the learning progression. Revisions to each path are made through a majority voting mechanism among the agents. 


For the research trajectory, we assess the plausibility and potential of the direction by evaluating the novelty of the LLM-generated Hint Idea $H$. Given that evaluating the originality of a research idea requires in-depth analysis and nuanced reasoning, we adopt a structured, step-by-step evaluation process. Specifically, we decompose the assessment of idea novelty into three structured stages, and develop a dedicated evaluation model, \textbf{GoAI-CoT-Reviewer}, to perform this task through chain-of-thought reasoning \cite{weiChainofThoughtPromptingElicits2023}, enabling a more rigorous and interpretable judgment of innovation.

\subsubsection{Step-by-Step Evaluation Stages}
\begin{figure}[!t]
    \centering
    \includegraphics[width=\linewidth]{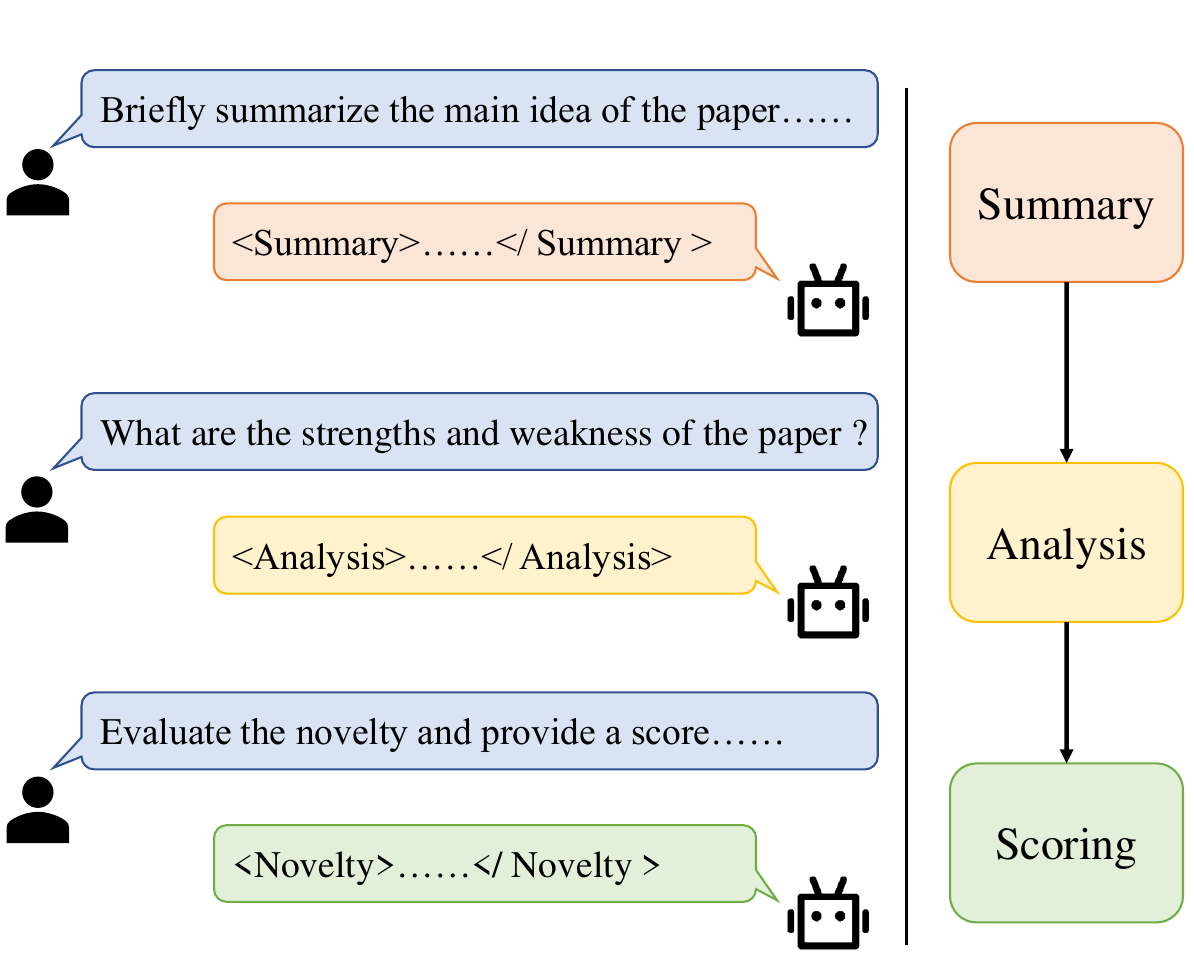}
    \caption{Processing flow of novelty evaluation for GoAI-CoT-Reviewer.}
    \label{fig:evaluation-framework}
\end{figure}


As shown in Figure \ref{fig:evaluation-framework}, incorporating the thought process of human reviewers, we decompose the evaluation of paper novelty into three structured stages: summary, analysis, and scoring.

\textbf{Summary Stage}: In this initial stage, GoAI-CoT-Reviewer is provided with an idea and a corresponding paper abstract based on this idea. GoAI-CoT-Reviewer briefly summarizes the main content and methods expressed by the idea based on this information.

\textbf{Analysis Stage}: Based on the summarized content and methods, GoAI-CoT-Reviewer analyzes the strengths and weaknesses of the idea, which will serve as the basis for the subsequent scoring.

\textbf{Scoring Stage}: Based on the results obtained from the first two stages, GoAI-CoT-Reviewer provides a numerical score to assess the novelty of the idea.


GoAI-CoT-Reviewer uses dedicated tags (e.g., \textit{$<$Summary$>$...$<$/Summary$>$}) to mark each stage, indicating the beginning and end of each phase. These tags help the model maintain clarity throughout the reasoning process. Unlike traditional CoT reasoning, which allows the model to freely think, our approach promotes structured thinking by first summarizing the input information, then performing a detailed analysis, and finally arriving at a score. To achieve this, we construct a training dataset by progressively generating responses using open review comments from the OpenReview platform and then train the evaluation model using supervised fine-tuning. After training, the model autonomously selects these labels as needed, completing all stages within a single process. This structured approach enables the model to independently manage its reasoning process, improving its performance on complex tasks that require multi-step analysis.

\subsubsection{Data Preparation and Model Training}
We use the Openreview API v2\footnote{https://docs.openreview.net/reference/api-v2} to scrape open-source AI conference papers and corresponding review comments submitted on the Openreview platform, including both accepted and rejected papers. To ensure the quality of the papers and review comments to some extent, we scraped papers and reviews from top AI conferences such as ICLR 2022-2025, NeurIPS 2022-2025, with a total of over 38k samples in the training dataset.

We use the \textit{summary of the paper} field from the review comments as the golden answer for the Summary stage, the \textit{summary of the paper} and \textit{strengths and weaknesses} fields as the golden answers for the Analysis stage, and the \textit{technical novelty and significance} field as the golden answer for the Scoring stage to construct a supervised fine-tuning dataset. We fine-tune the base model using LoRA to obtain the GoAI-CoT-Reviewer model, which is used for path validation.

\subsubsection{Multi-agent Evaluation}

After training the GoAI-CoT-Reviewer, we assemble a multi-agent evaluation system comprising multiple GoAI-CoT-Reviewer agents to assess the novelty of each Hint Idea, thereby providing a reference for determining the plausibility and value of the corresponding research trajectory. Similar to the validation process for learning paths, each GoAI-CoT-Reviewer agent is prompted to generate a qualitative assessment and assign a numerical score ranging from 1 to 10. A threshold score of 5 is employed to distinguish between promising and unpromising ideas, aligning with the acceptance criteria used in peer review. The final judgment on the quality of an idea is then determined through a majority vote among agents.

\begin{table*}[!t]
\centering
\small
\begin{tabular}{@{}c|c|ccccccc|c@{}}
\toprule
\multirow{2}{*}{} & \multirow{2}{*}{SUS $\uparrow$}  & \multicolumn{7}{c|}{NASA-TLX $\downarrow$}  & \multirow{2}{*}{Avg. Idea Score $\uparrow$}  \\ \cmidrule(lr){3-9}
   &                  & Mental & Physical & Temporal & Performance & Effort & Frustration & Avg.  &      \\ \midrule
Group 1 (Web) & \textbackslash{} & 15.33  & 7.00     & 16.67    & 13.67       & 17.00  & 13.33       & 13.83 & 5.67 \\
Group 2 (LLM) & 85.33            & 8.33   & 7.67     & 7.67     & 8.67        & 6.67   & \textbf{7.33}        & 7.72  & 6.33 \\
Group 3 (GoAI) & \textbf{87.67}            & \textbf{5.33}   & \textbf{6.33}     & \textbf{3.67}     & \textbf{8.67}        & \textbf{4.67}   & 7.67        & \textbf{6.06}  & \textbf{7.33} \\ \bottomrule
\end{tabular}%
\caption{The performance of student groups employing different methods.}
\label{tab:main}
\end{table*}

\begin{table}[!t]
\centering
\small
\begin{tabular}{c|c}
\hline
Method                     & Pearson Corr. \\ \hline
Direct Generation          & 0.0058 \\ 
Multi-turn Dialogue        & 0.4251 \\ 
Structured Thinking (ours) & 0.6810 \\ \hline
\end{tabular}
\caption{Pearson correlation coefficients between different
evaluation models and human assessment.}
\label{table:sft-dataset}
\end{table}

\subsection{Idea Studio}

We present \textbf{Idea Studio} as a demonstrative application of the GoAI framework. For an AI major student, Idea Studio provides structured prompts requiring them to conduct research and formulate novel ideas based on a specific topic—for instance, ``Enhancing the Reasoning Ability of Large Language Models.” By leveraging GoAI to generate research trajectories and personalized learning paths, students gain access to the developmental trends and key literature within the selected domain, enabling them to independently acquire the necessary prerequisite knowledge. Upon completing the study of relevant concepts and reviewing the associated papers, students propose original ideas informed by the identified trends, which are then evaluated and iteratively refined through the GoAI-CoT-Reviewer. This process not only serves as an empirical evaluation of GoAI’s practical utility but also functions as a pedagogical tool to support the everyday learning and research activities of AI students.

\section{Experiments}

\subsection{Setup}

We use the Deepseek-V3 API as the backend LLM. The GoAI-CoT-Reviewer evaluation model is built on the Llama 3.1-8B-Instruct model \cite{dubey2024llama} and fine-tuned through LoRA. Training is conducted using NVIDIA A100 GPUs, with a LoRA rank set to 8.

\subsection{Evaluation Method}

As the only trainable component within the GoAI system, we first conducted an experimental evaluation of the correlation between the GoAI-CoT-Reviewer and human judgments, thereby establishing the reliability and practical utility of the GoAI framework as a whole and laying the groundwork for subsequent experiments.

 To evaluate the overall effectiveness of the GoAI system, we recruited 9 upper-level undergraduate and early-stage graduate students majoring in AI, all possessing foundational domain knowledge, and assigned them to one of three groups in a between-subjects randomized design. Each group employed a distinct method to analyze research trends and generate ideas: (G1) web search combined with manual paper reading, (G2) interaction with an LLM-based tutoring assistant, and (G3) use GoAI Idea Studio. Participants were instructed to complete two tasks within a two-hour time frame: Frontier Mapping, which involved drafting a one-page summary including key baselines, extensions, contrasts, datasets/tools, and a list of prerequisite knowledge; and Creative Ideation, which required proposing one to two project ideas related to the assigned topic. Participants using LLM and GoAI were asked to complete the System Usability Scale (SUS), while all participants completed the NASA-TLX to assess task load. Additionally, we recorded the scores of the ideas generated by all participants.

\subsection{Main Results}

\subsubsection{Human-Model Agreements of Evaluation}

Table \ref{table:sft-dataset} presents the Pearson correlation coefficients between our evaluation model and human assessments. The results indicate that our model demonstrates a strong positive correlation with human reviewers, validating its effectiveness as a source of feedback in the idea generation process. Compared to models trained on raw data or multi-turn dialogue datasets, those incorporating a structured Chain-of-Thought (CoT) reasoning process exhibit significantly improved alignment with human evaluations, a finding that is consistent with the conclusions drawn in \cite{xuLLaVACoTLetVision2025}.

\subsubsection{The Effectiveness of GoAI}

Table \ref{tab:main} presents the performance of student groups employing different methods, highlighting the effectiveness of GoAI. In terms of system usability, the scores for GoAI and the LLM-based assistant are comparable, indicating that despite the greater complexity of the GoAI system, it does not impose additional operational burdens or steep learning curves on users. Regarding task-related cognitive load, students using GoAI reported lower levels of mental and time pressure, suggesting that the GoAI approach effectively supports AI students in synthesizing research trends, identifying prerequisite knowledge, and generating novel research ideas. Notably, students utilizing GoAI also outperformed their peers in the novelty of the ideas produced, demonstrating its capacity to enhance creative ideation in academic contexts.

\section{Conclusion}

In this paper, we present Graph of AI Ideas (GoAI), a novel framework that integrates LLMs with knowledge graphs to support AI students in systematically understanding research trends, constructing prerequisite-aligned learning pathways, and generating innovative research ideas. By organizing ideas, methodologies, and conceptual elements from existing literature into a structured graph representation, GoAI effectively captures the progressive relationships among research works. Performing path search within this graph enables the LLM to better comprehend and analyze ongoing advancements, uncover interconnections among studies, and thereby generate more coherent developmental trajectories and tailored learning paths. To evaluate and iteratively refine the automatically generated ideas, we introduce GoAI-CoT-Reviewer, a structured reasoning-based evaluation model that assesses the novelty of research ideas through a process closely aligned with human peer review, yielding feedback with high agreement to expert judgment. Experimental results demonstrate that GoAI offers strong usability, significantly reduces students’ cognitive load in synthesizing research landscapes and learning requirements, and effectively fosters the generation of higher-quality and more creative research ideas.

\clearpage
\bibliography{aaai2026}

\clearpage

\end{document}